\documentclass[10pt,twocolumn]{article}

\usepackage{cvpr}
\usepackage{times}
\usepackage{epsfig}
\usepackage{graphicx}
\usepackage{amsmath}
\usepackage{amssymb}
\usepackage{multirow}

\usepackage[pagebackref=true,breaklinks=true,letterpaper=true,colorlinks,bookmarks=false]{hyperref}

\cvprfinalcopy 

\def\cvprPaperID{1208} 

\ifcvprfinal\fi
\begin{document}

\newcommand{\todo}[1]{\textcolor[rgb]{1,0,0}{#1}}
\newcommand{\keypoint}[1]{\vspace{0.0cm}\noindent\textbf{#1}\quad}

\title{The Devil is in the Middle: Exploiting Mid-level Representations for Cross-Domain Instance Matching}

\author{Qian Yu$^1$  \quad \quad Xiaobin Chang$^1$ \\
Yi-Zhe Song$^1$ \quad\quad Tao Xiang$^1$ \quad \quad Timothy M. Hospedales$^2$ \\
Queen Mary University of London, London, UK$^1$ \\
University of Edinburgh, Edinburgh, UK$^2$ \\
{\tt\small \{q.yu, x.chang, yizhe.song, t.xiang\}@qmul.ac.uk \quad\quad \tt\small t.hospedales@ed.ac.uk}
}

\maketitle
\thispagestyle{empty}

\begin{abstract}
Many vision problems require matching images of object instances across different domains. These include fine-grained sketch-based image retrieval (FG-SBIR) and Person Re-identification (person ReID). Existing approaches attempt to learn a joint embedding space where images from different domains can be directly compared. In most cases, this space is defined by the output of the final layer of a deep neural network (DNN), which primarily contains features of a high semantic level. In this paper, we argue that both high and mid-level features are relevant for cross-domain instance matching (CDIM). Importantly, mid-level features already exist in earlier layers of the DNN. They just need to be extracted, represented, and fused properly with the final layer.  Based on this simple but powerful idea, we propose a unified framework for CDIM. Instantiating our framework for FG-SBIR and ReID, we show that our simple models can easily beat the state-of-the-art models, which are often equipped with much more elaborate architectures.

\end{abstract}

\vspace{-0.3cm}
\section{Introduction}


A number of widely studied computer vision problems can be seen as cross-domain instance matching (CDIM) problems. In these problems,  object instances are captured in multiple domains; and given a query item in one domain, the goal is to find among a gallery set the correct matching items that contain the same object instance from the other domain. These problems differ mainly in the  domains where an object instance is captured. For example, in face verification \cite{schroff2015facenet,hu2017ntfFace} or person re-identification \cite{yi2014deep, sun2014deep,ahmed2015improved}, the object instances (person face or full body) are captured by different camera views; in instance-level fine-grained sketch-based image retrieval (FG-SBIR) \cite{yi2014bmvc,yu2016sketch,sangkloy2016sketchy}, the domains are the photo and sketch modalities.  CDIM problems can also generalise beyond images, \eg, in captioning where the domains are images and text \cite{karpathy2015deep, vinyals2015show}; but in this work we focus on CDIM problems where both object instances are represented as images.

\begin{figure}[t!]
\centering
\includegraphics[width=1\linewidth]{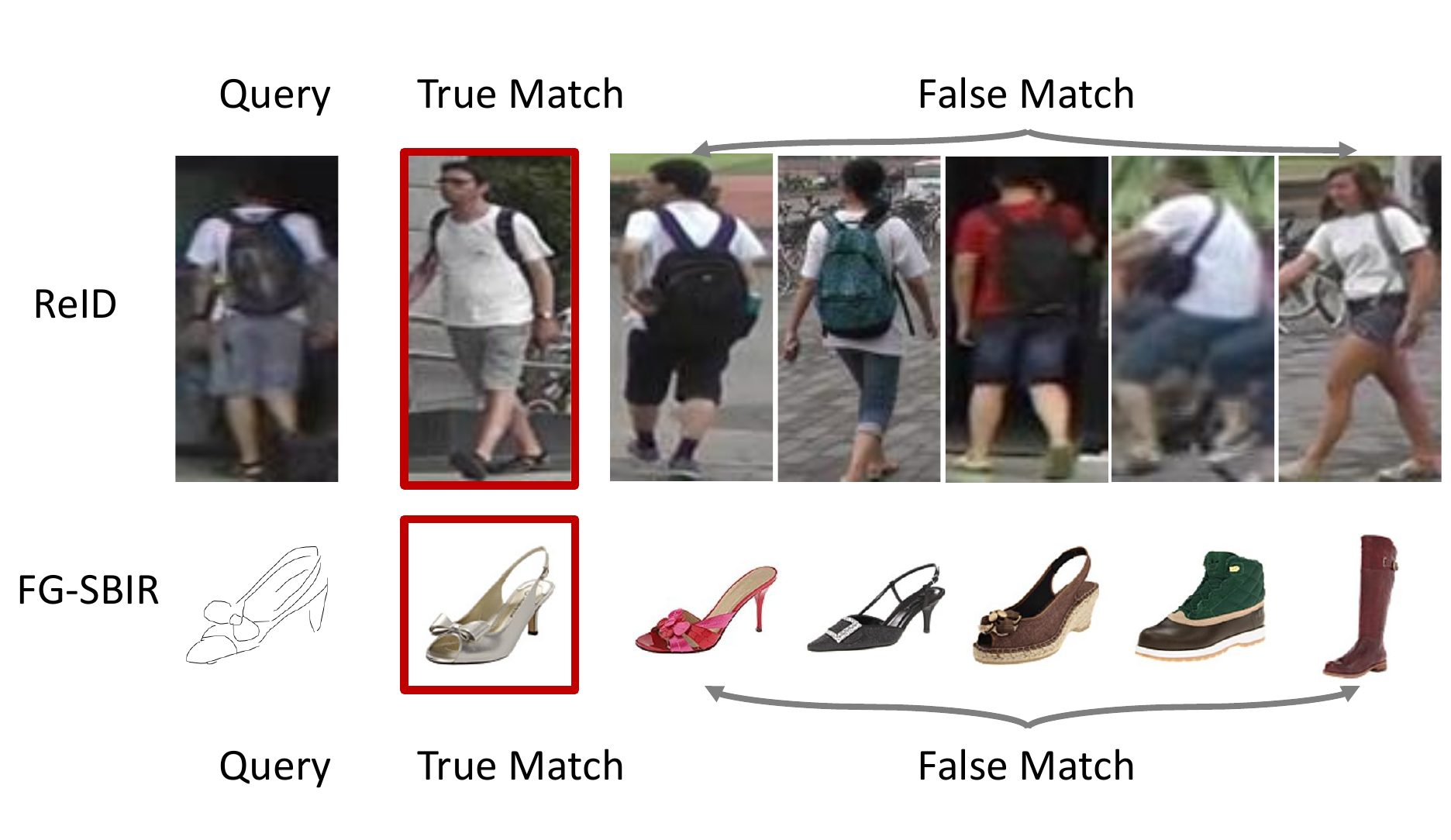}
\vspace{-0.5cm}
\caption{A CDIM task is very challenging. Top: Person ReID; Bottom: FG-SBIR.}
\label{fig:motivation}
\end{figure}


A CDIM problem faces a number of challenges. Taking the person ReID and FG-SBIR problems as examples, these challenges are illustrated in Fig.~\ref{fig:motivation}: (1) The existence of domain gap means that the visual appearance of the same instance can look very different. For instance, in FG-SBIR, color and texture information is missing from sketch; and free-hand sketches feature distorted object contours and a variable level of abstraction that differs significantly from the matching photo image. In ReID, this challenge takes a different form: the person's body pose, image background, lighting, and viewpoints change drastically across camera views. (2) The discriminative features that can distinguish different object instances are often subtle with a small spatial support  in the image. This is because all instances are from the same object category (\eg, person, shoe) and thus share many visual similarities. The differences are often found in small regions such as the buckle on the shoe and the bag carried by the person. (3) These discriminative features correspond to visual concepts of different semantic levels. For instance, in the shoe FG-SBIR example, the high-level sub-category concept (\eg, heels vs.~trainers) can help narrow down the search, but mid-level concepts such as the bowknot on the heels  pinpoint the exact instance. Similarly, in the ReID case, gender  narrows down the search scope but clothing colour at different regions (top, bottom, shoe) does the final matching.  Considering these challenges, an effective CDIM model should be able to extract features that are domain-invariant, discriminative at different semantic levels and spatially sensitive.  

\begin{figure*}[!t]
\centering
\includegraphics[width=\linewidth]{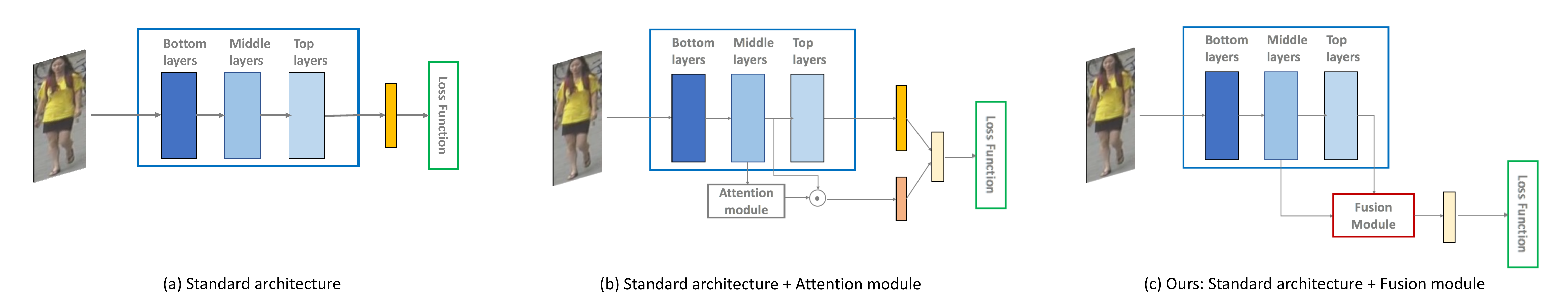}
\vspace{-0.5cm}
\caption{Our proposed architecture. Blue box: the CNN module. Red box: the Fusion module. Green box: the Loss module.  The base CNN can be changed to any architecture.}
\label{fig:architecture}
\end{figure*}

Most state-of-the-art CDIM models \cite{yi2014deep, sun2014deep,ahmed2015improved,yu2016sketch}  use deep neural networks (DNNs) to learn a joint embedding for different domains. The hope is that in this space the domain gap is removed and  object instance images from different domains can be directly compared, \eg, by simple Euclidean distance. They thus mostly focus on solving the first challenge, \ie the appearance difference caused by the domain gap. The other two challenges of salient cues requiring spatial support and multiple semantic levels are largely ignored. In particular, the embedding space is typically defined by the features extracted from the final feature layer of a DNN, after global pooling (see Fig.~\ref{fig:architecture}(a)). This feature extraction strategy worked well for object category recognition, generating super-human performance on the ImageNet 1K classification task \cite{ILSVRC15}. However, it is less suitable for the CDIM problems studied here: The loss of spatial support due to global pooling is detrimental for instance matching. Meanwhile it is widely acknowledged~\cite{DL_nature} that, progressing from  bottom to  top layers, DNN feature maps tend to correspond to visual concepts that are more abstract and of higher semantic level. However, as mentioned earlier, the discriminative features for CDIM need to capture visual concepts of both high and middle semantic levels.


More recently, the problem of CDIM embedding space containing only high-level visual concepts has been identified for both FG-SBIR \cite{song2017sketch} and person ReID \cite{liu2017hydraplus}. Their solutions are based on soft-attention (see Fig.~\ref{fig:architecture}(b)). Specifically, attention maps are computed from middle layers of a DNN and used to weight the corresponding feature maps and fused with the final layer feature output as the final representation/joint embedding space. However, the attended feature maps are still globally pooled thus losing all spatial information. More importantly the attention maps introduce additional parameters and are often very noisy \cite{song2017sketch} for the fine-grained instance recognition problems.


\begin{figure}[!t]
\centering
\includegraphics[width=1\linewidth]{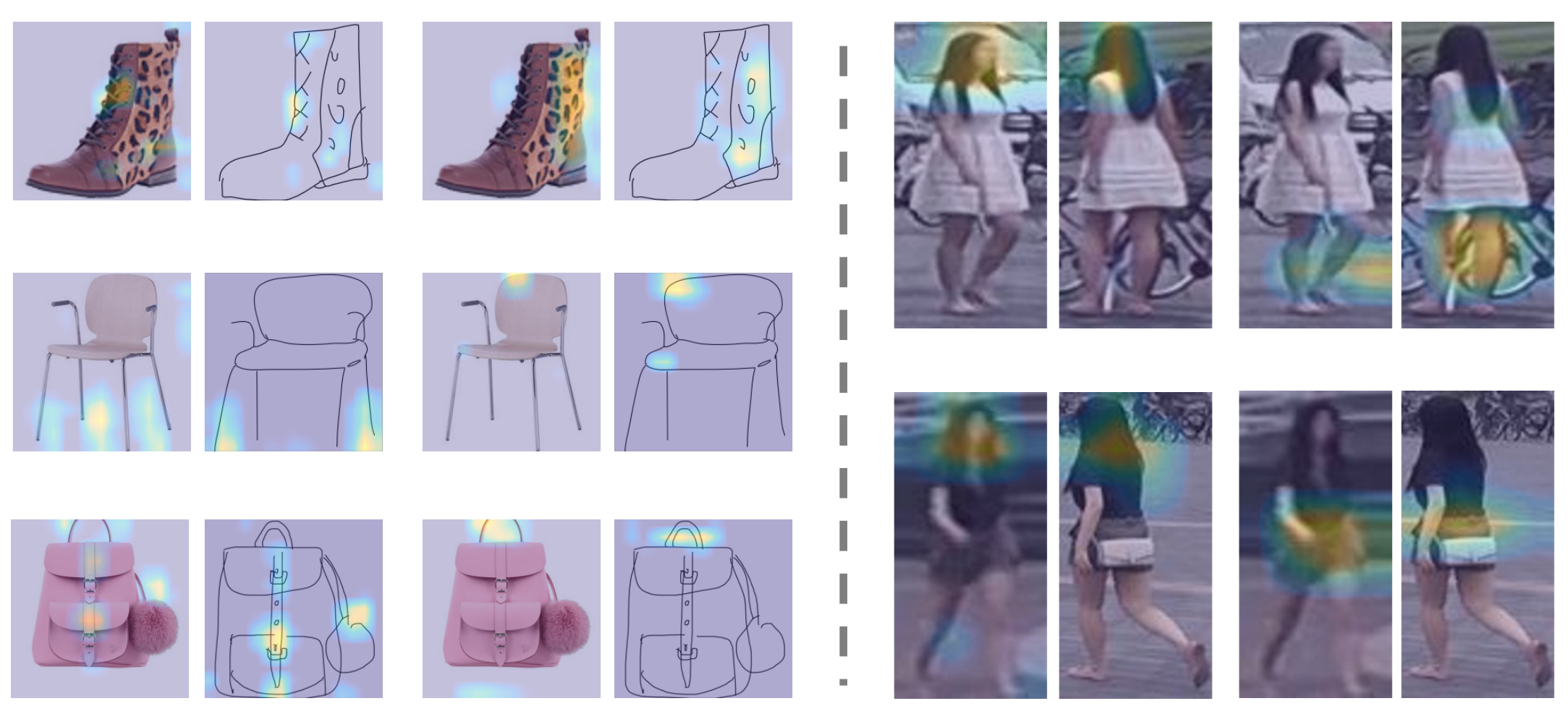}
\vspace{-0.5cm}
\caption{\textcolor{black}{Visualizations of the feature map activated by different filters of layer \textit{conv5} of Sketch-a-Net \cite{qianyu2015bmvc} when modeling FG-SBIR data on the left,  and \textit{res5b} of ResNet-50 \cite{he2016deep} when modeling ReID data on the right. They capture the mid-level concepts like `shoelace', `chair legs', `bag handle' and `head' etc.,  which are discriminative for cross-domain matching.
}}
\label{fig:filters}
\end{figure}

In this paper, we observe that spatially-sensitive and mid-level features are already there -- the feature maps computed from the mid-layers of a DNN trained for CDIM capture exactly that (see Fig.~\ref{fig:filters} for some examples).  So no elaborate architectural modification such as introducing attention modules as in \cite{song2017sketch,liu2017hydraplus} is required. One only needs to (1) provide deep supervision to make them more discriminative for the CDIM task at hand, and (2) design an appropriate feature representation based on the feature maps and a fusion module to fuse them with the final-layer high-level features. The proposed architecture is shown in Fig.~\ref{fig:architecture}(c). We present insights supported by extensive experiments on how to design the mid-level representation and the fusion module. Based on the unified architecture design and the insights, we propose two models for FG-SBIR and person ReID respectively. We show that with a simple model design, the two models comfortably beat  the existing state-of-the-art models, many of which have very elaborative and often expensive architectures. 

Our contributions are: (1) For the first time we observe that the mid-layer feature maps produced by a CDIM DNN contain  the missing mid-level domain-invariant discriminative features required for effective CDIM. (2) We propose a unified CDIM DNN architecture design pattern to take advantage of this  finding. (3) Two instantiations of the proposed framework achieve new state-of-the-art results on three FG-SBIR and three person ReID benchmark datasets.  

\vspace{-0.2cm}
\section{Related Work}

\keypoint{Fine-grained SBIR} With the growth of touch-screen devices, the problem of fine-grained image retrieval (FG-SBIR) has received increasing interest recently  \cite{yi2014bmvc,yu2016sketch,sangkloy2016sketchy,
song2017sketch}.  It was first proposed in \cite{yi2014bmvc} where a sketch is used to retrieve a gallery photo containing an object in the same pose or viewpoint, rather than the same \textcolor{black}{object instance per se}. The instance-level FG-SBIR problem was first formalized and addressed in \cite{yu2016sketch} targeting an online shopping application, \eg a user sketches a specific shoe to search for it among all the shoes available in an online shopping website. In this scenario, the pose or viewpoint is aligned between the query sketch and the gallery photos and the gallery contains only object photos of the same category as the query sketch.  Differently \cite{sangkloy2016sketchy} studies FG-SBIR in the presence of objects of multiple categories. Both works employ a deep triplet ranking model to learn a domain-invariant  representation shared by both the photo and sketch domains. Only the final CNN layers are used as the representation; thus primarily high-level semantic features are included. 

This problem is alleviated in \cite{song2017sketch} through an attention module that takes a mid-layer as input. We show here that the attention module is redundant -- the mid-layer feature map alone is complementary to the final output and can be directly fused with it. We notice that the feature map encodes discriminative spatial information about where salient features are located in the image. However, such information is lost using the attention module in \cite{song2017sketch}, limiting its effectiveness. We show in our experiments that our simpler model outperforms that of \cite{song2017sketch} by a clear margin. 
  
\keypoint{Person Re-identification} Person Re-identification (ReID) aims to match people across non-overlapping camera views in a surveillance camera network. Recent ReID models are mostly based on DNNs \cite{geng2016deep,zheng2016discriminatively,xiao2016learning,li2017person,hermans2017defense},  with similar multi-branch architectures as those used in FG-SBIR. Apart from triplet ranking loss, other losses such as identity classification loss and pairwise verification loss are also popular.  Different from FG-SBIR, person images captured by different cameras often have very different view-angles, body poses and occlusions. As a result, spatial information about where the discriminative features come from in the image is less useful in ReID. 

For deep ReID, learning discriminative view-invariant features from multiple semantic levels becomes even more important: low-level semantic concepts like color and texture are as important as  high-level ones such as carrying condition and gender (see Fig.~\ref{fig:motivation}). However, most existing deep ReID models use a DNN originally designed for object categorization and thus exploit the mainly  high-level features in the final-layer output. This drawback has been acknowledged recently and a number of works attempted to rectify it.  One line of work \cite{lin2017improving} resorts to additional supervision in the form of attributes \cite{layne2012person}. Attributes correspond to visual concepts of multiple semantic levels. By forcing the ReID model to also predict the attributes, it is hoped that the final-layer feature output can be encouraged to capture multi-level visual concepts. Some other works use a body pose model to detect local body parts and extract features from each part accordingly \cite{zhao2017spindle}. This can be considered as learning to model part-based attributes.   However, attribute based deep ReID models require additional manual annotation of attributes and cannot completely solve the problem of extracting multi-level features from a single layer. Instead we rely on the observation that deep networks already learn to represent attributes internally \cite{shankar2015deepCarving,vittayakorn2016neuralAttrDiscovery}, and focus instead on how to extract and exploit these learned attributes.

The most related work is  \cite{liu2017hydraplus}, which also attempts to extract multi-level discriminative features. However, their approach is based on computing soft attention maps at different middle layers of a network. The attended feature maps are still fed into the later layers and only the final layer is used for the final feature representation. In contrast, our model is much simpler and directly fuses the mid-layer feature maps with the final-layer one. Our experiments show that our model is much more effective despite its simplicity.  

\keypoint{Multi-level Fusion in DNNs}\quad Beyond CDIM, the idea of fusing multi-level/scale features learned at different layers has been exploited in other vision problems including semantic segmentation and edge detection estimation \cite{hariharan2015hypercolumns,xie2015holistically}. Similar to our proposed model, feature maps computed at different network layers are fused. However, in this case the purpose is mainly to incorporate multiple spatial granularities of information in image-to-image synthesis problems. In our case we aim to extract  mid-level localized salient latent attributes from selected middle layers in order to obtain a richer representation for cross-domain image matching.


\section{Methodology}


\subsection{Overview}\label{sec:method}
\begin{figure}[!t]
\centering
\includegraphics[width=1\linewidth]{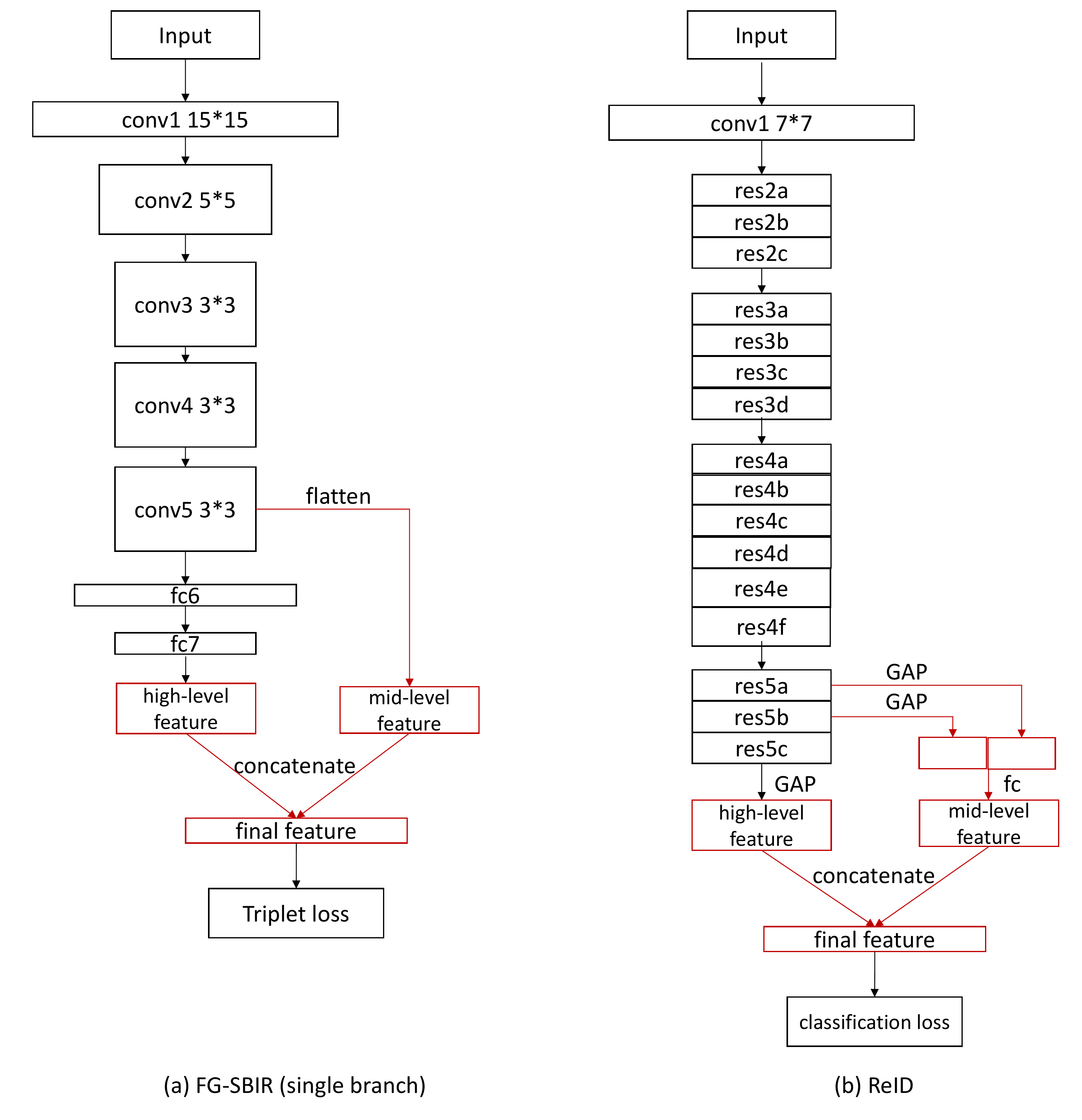}
\vspace{-0.5cm}
\caption{Proposed model for each case study. (a) A single branch of the FG-SBIR model. (b) The ReID model. The fusion module is highlighted in red.}
\label{fig:structure}
\end{figure}

Our proposed design-pattern is to learn a feature representation containing both the high-level and mid-level semantic information which are beneficial for CDIM tasks. The proposed design pattern is illustrated in Fig.~\ref{fig:architecture}(c). It consists of three parts: (1) a CNN base network (blue box); (2) a fusion module (red box) and (3) a loss module (green box). The key component of the framework is the fusion module. Specifically, we take feature maps of one or more mid-layer(s) of the base network and construct a feature vector by applying an appropriate pooling method. The mid-level feature and the final-layer feature of the CNN base network are then be concatenated to form the final feature representation, which will be subjected to the loss module during training. This  strategy both ensures that mid-level details are included in the representation and that the mid-level details are trained to be discriminative via deep supervision. The CNN base network can be any existing CNN structure, \ie ResNet-50, InceptionV3, etc. Based on this framework, we propose two specific instantiations for FG-SBIR and person ReID respectively (as shown in Fig.~\ref{fig:structure}). 

\subsection{FG-SBIR Model}

\keypoint{CNN Module}\quad We exploit the widely used Siamese triplet architecture \cite{yu2016sketch,bui2016generalisation,song2017sketch}. It has three branches with all parameters shared. The three branches take a query sketch, a positive photo and a negative photo as  input respectively. For the single branch $F$, we adopt the Sketch-a-Net \cite{qianyu2015bmvc} as the base network which was originally proposed for sketch recognition. It includes 5 convolutional layers and 2 fully-connected layers. Given an image, the final-layer (\textit{fc7}) output is $\vec{f}^{fc7} \in \mathbb{R}^{C}$ where $C$ is the number of feature channels. Instead of being used as the image representation directly, this feature, along with the feature maps of the fifth convolutional/pooling layer $\vec{f}^{conv5} \in \mathbb{R}^{H \times W \times C}$ ($H$ and $W$ are the size of the feature maps) will be fed into the fusion module for further processing. Figure \ref{fig:filters} shows several visualisations of the feature map in $\vec{f}^{conv5}$.

\keypoint{Fusion Module}\quad Based on the features of the middle and final layers of the base network, the fusion module will generate the final feature representation. The first thing is to transform the mid-level feature vector (\ie from $\vec{f}^{conv5} \in \mathbb{R}^{H \times W \times C}$ to $\vec{f}^{mid} \in \mathbb{R}^{C^*}$, $C^*$ is the number of activations of the feature vector). An appropriate transformation method needs to be selected. The most common way is  global average pooling (GAP); however, spatial information will be lost. Given that sketches and photos are already pose-aligned,  spatial information will be useful and should be kept in the final feature representation. So we flatten the $\vec{f}^{conv5}$ directly. Our experiment results verify the benefit of  retaining spatial information with flattening  instead of discarding it with GAP (Sec.~\ref{sec:discuss}). The mid-level feature vector  can be represented as 
\begin{equation}
\vec{f}^{mid} = \text{flatten}(\vec{f}^{conv5})
\end{equation}
Next, this mid-level feature $\vec{f}^{mid}$ and the final-layer feature $\vec{f}^{fc7}$ are L2 normalized respectively and then  concatenated to form the final feature representation $\vec{f}^{final}$. A fully connected layer could be added after $\vec{f}^{mid}$ to reduce the dimension. However, this would introduce more parameters and loss of spatial information. So we did not further reduce the  dimension of $\vec{f}^{mid}$. 

\keypoint{Loss Module} The output of the fusion module will be fed into the loss module during training. A triplet ranking loss is used for our triplet architecture. For a given triplet tuple $t=(q,p^+,p^-)$, this loss is defined as:
\begin{equation}
L_{\theta}(t) = \max (0, \Delta + D(F_{\theta}(q),F_{\theta}(p^+)) - D(F_{\theta}(q),F_{\theta}(p^-)) )
\label{rankLoss}
\end{equation}
where $\theta$ are the parameters of the CNN with the fusion module, $F_\theta(\cdot)$ denotes the output of the corresponding branch, \ie $\vec{f}^{final}$, $\Delta$ represents a margin between the query-positive distance $D(F_{\theta}(q),F_{\theta}(p^+)$) and query-negative distance $D(F_{\theta}(q),F_{\theta}(p^-))$ ($\Delta$ is 0.3 in our implementation). Euclidean distance is used here.

\subsection{Person ReID Model}

\keypoint{CNN Module}\quad For ReID, we adopt a classification model with ResNet-50 \cite{he2016deep} as the base network. It has no fully connected layers -- all parameter layers are convolutional layers. It is a common practice to use the last convolutional layer feature (\textit{res5c}) after  global average pooling as the final image representation, denoted as $\vec{f}^{pool5}$. However, it only contains higher-level semantic information and we are also interested in learning discriminative mid-level semantic features. Therefore we select feature maps of \textit{res5a} and \textit{res5b}, denoted $\vec{f}^{res5a}$ and $\vec{f}^{res5b}$, as the mid-level features to complement the $\vec{f}^{pool5}$. They will be subjected to the fusion module to generate the final feature representation. Figure \ref{fig:filters} shows several visualizations of the feature map in $\vec{f}^{res5b}$.

\keypoint{Fusion Module}\quad Different to FG-SBIR, pose-variation is a big problem for ReID -- as shown in Fig.~\ref{fig:motivation}, the same identity presents significantly different poses in different camera views, which suggests that  the spatial information is not  reliable any more. We thus apply a different pooling strategy to generate the mid-level feature vector.  Specifically, when generating the mid-level feature vector $\vec{f}^{mid}$, we do global average pooling (GAP) instead of flattening on $\vec{f}^{res5a}$ and $\vec{f}^{res5b}$. After that, these two vectors are concatenated, followed by a fully connected layer to reduce the dimension. 
Finally, the mid-level feature vector is concatenated with the top-layer output $\vec{f}^{pool5}$ as the final image representation $\vec{f}^{final}$.

\keypoint{Loss Module:}\quad During training, the final image representation $\vec{f}^{final}$ will be fed into a softmax cross entropy loss to predict the training identities. Given an image $p$, 
\begin{equation}
\label{equ:clf_loss}
L_{clf}(p) = \sum_{k=1}^{K} {-f_k(p)log(\hat{f}_k(p))},
\end{equation}
where $K$ is the class number (the number of the identities), and $\hat{f}_k(q)$ is the softmax output of the logits layer.

\subsection{Alternatives}
\label{sec:alternatives}

Our fusion strategy is to combine  mid and  high-level representations to make the final feature representation contain semantic information at different levels. However, there are two questions when we develop this module: (1) The feature maps of which layer(s) should be selected for fusion? (2) What pooling strategy should be used to generate the mid-level feature vector? These two factors are important in determining the effect of the fusion module.

\keypoint{Fuse All Layers vs. Fuse Specific Layer(s)}\quad As mentioned before, the earlier layer(s) capture low-level concepts like color and edges while the deeper layer(s) extract more semantic and abstract concepts \cite{vittayakorn2016neuralAttrDiscovery}. We aim to capture the mid-level representation corresponding to semantic part information. Such features are less abstract -- more concrete and local --  than the features generated by the top layer. It is intuitive to look the layers slightly lower than the final layer. Recent studies  \cite{bau2017network,vittayakorn2016neuralAttrDiscovery,diba2016deepCamp} discovered that filters at \textit{conv5} of an AlexNet (Sketch-a-Net is a variant of AlexNet) represent concrete attributes such as nameable pieces of clothing \cite{vittayakorn2016neuralAttrDiscovery} and localizable parts \cite{diba2016deepCamp}. 
Although ResNet-50 is a much deeper network, it consist of 5 groups of convolutional layers and each group is analogous to a convolutional layer in AlexNet style architectures like Sketch-a-Net. Given these observations, we use the activations of \textit{conv5} (of the Sketch-a-Net) and \textit{res5a} and \textit{res5b} (of the ResNet-50) as the mid-level representation for fusion. 
We do experiments to compare fusing lower layers like \textit{conv4} or \textit{res4x}, and even the extreme case of fusing all layers. The results in \textcolor{black}{Sec.~\ref{sec:discuss}} show that fusing the mid-level representation \textit{conv5} (or \textit{res5a} and \textit{res5b}) achieves the optimal performance as expected from prior insights \cite{bau2017network,vittayakorn2016neuralAttrDiscovery,diba2016deepCamp}. 

\keypoint{Task-specific Pooling Method}
To generate the mid-level feature vector, different pooling/flattening strategies are applied in FG-SBIR and ReID due to their different properties with regards to spatial alignment. Specifically, in FG-SBIR, we flatten  $\vec{f}^{conv5}$ directly in order to keep the spatial information. While in ReID, we do global average pooling on $\vec{f}^{res5a}$ and $\vec{f}^{res5b}$ to get rid of the spatial information given the large pose-variation across views. We will show that this task-specific pooling strategy is necessary by comparing them in both tasks. The results are provided in Sec.~\ref{sec:discuss}, Table~\ref{tab:pooling FG-SBIR} and Table~\ref{tab:pooling reid}. 



\section{Experiments and Results}
\label{Sec:exp}

\subsection{Experiment Settings}

\begin{table}[!t]
\normalsize
\centering
\resizebox{0.75\columnwidth}{!}{
\begin{tabular}{c|c|c|c}
\hline    
\multirow{2}{*}{Item} & \multicolumn{3}{|c}{QMUL FG-SBIR}\\
\cline{2-4}
& Shoe & Chair & Handbag \\                            
\hline
\#Photos & 419 & 297 & 568  \\
\hline
\#Sketches &  419 & 297 & 568 \\
\hline
\#Triplet annotations& 13,689 & 9,000 & -- \\
\hline
\#Sketches/Photos (train) & 304/304 & 200/200 & 400/400 \\
\hline
\#Sketches/Photos (test) & 115/115 & 97/97 & 168/168 \\
\hline
\end{tabular}}
\vspace{0.2cm}
\protect\caption{The training/testing splits of each category in the QMUL FG-SBIR database.}
\centering
\label{tab:FG-SBIR}
\end{table}

\keypoint{Dataset Splits and Pre-processing}
\keypoint{FG-SBIR} We use the QMUL Fine-grained sketch database \cite{yu2016sketch,song2017sketch} to evaluate our proposed method on FG-SBIR. This database includes three categories, shoes, chairs and handbags, and each of them serves as an independent benchmark. The photos were collected from online shopping websites and the free-hand sketches for each photo were collected via crowd-sourcing. Apart from photo-sketch pairing, additional annotations are available for shoes and chairs in the form of triplet ranking consisting of one sketch and two photos ranked based on similarity to the sketch. The details of training/testing splits are listed in Table~\ref{tab:FG-SBIR}. During training the models for shoes or chairs, following \cite{yu2016sketch,song2017sketch}, human triplet annotations are used to generate positive and negative photos given a query sketch. For handbag, the true match photo is used to form positive pair  while any false match is used for the negative pair. For data pre-processing, we follow the same procedure described in \cite{yu2016sketch}, including the staged pre-training.

\keypoint{Person ReID}\quad Three large benchmark datasets, Market-1501 \cite{zheng2015scalable}, DukeMTMC\_reID \cite{zheng2017unlabeled} (a subset of the DukeMTMC \cite{ristani2016MTMC}) and CUHK03-New (new setting introduced in \cite{zhong2017re}) are used to evaluate our proposed model. The statistics and training/testing splits of these three datasets are summarised in Table~\ref{tab:reid}.

\begin{table}[!h]
\normalsize
\centering
\resizebox{1.0\columnwidth}{!}{
\begin{tabular}{c|c|c|c}
\hline    
Item & Market-1501 & DukeMTMC\_reID & CUHK03-New \\                    
\hline
\#Cameras & 6 & 8 & 6 \\
\hline
\#Identities &  1,501 & 1,404 & 1,467  \\
\hline
\#Identities (training) & 751 & 702 & 767 \\
\hline
\#Identities (testing) & 750 & 702 & 700\\
\hline
\#Bounding box (training) & 12,936 & 16,522 & 7,368 \\
\hline
\#Bounding box (gallery) & 19,732 & 17,661 & 5,328 \\
\hline
\#Bounding box (probe) & 3,368 & 2,228 & 1,400 \\
\hline
\ Annotations & DPM & manually &DPM, manually \\
\hline
\end{tabular}}
\vspace{0.2cm}
\protect\caption{Training/testing splits of Person ReID datasets.}
\centering
\label{tab:reid}
\end{table}

\keypoint{Evaluation Metrics}\quad To measure the performance of the FG-SBIR and Person ReID, the first metric we use is the retrieval accuracy at top K, \eg acc.@K (`Rank-1' for ReID). This represents the percentage of the true match for a given query appearing in the top K. We also use the mean Average Precision (mAP) to evaluate the ReID performance. It is computed by averaging the area under the precision-recall curve over all queries.

\keypoint{Implementation Details}\quad
We use Tensorflow to implement our models. For \textbf{FG-SBIR}, we employ the Sketch-a-Net as base network, the same as \cite{yu2016sketch,song2017sketch}. For fair comparison, we follow the same experiment setting as described in \cite{yu2016sketch}, \ie we use SGD as optimizer, the mini-batch size is 128 and the initial learning rate is 0.001. We set the maximum training iterations as 20K but early-stopping is required due to overfitting. During training, images are randomly cropped to be 225$\times$225, and may be flipped with a probability 0.5. All models are fine-tuned on the step3 model released by \cite{yu2016sketch}. For \textbf{Person ReID}, ResNet-50 is used as the base network which provides a strong baseline. The optimizer is Adam and the learning rate decay factor and initial learning rate is set to be 0.95 and 0.00035 except CUHK03-New with 0.0005. We set the mini-batch size as 64 and the maximum training iterations as 60K and 75K for CUHK03-New (reduced from normal 100K to save time) . 
 All person ReID models are fine-tuned on ImageNet pre-trained model. The input image size is 256$\times$128 in order to keep a consistent human body ratio. Random flipping is applied during training.


\subsection{Results}

\keypoint{Results on FG-SBIR} Comparisons between ours and the state-of-the-art  are presented in Table \ref{tab:results_FG-SBIR}.  We  make  the following observations: (1) Overall, our proposed method outperforms all competitors over all three datasets, particularly at acc.@1. (2) To evaluate how much the introduced mid-level feature fusion contributes to our model's performance, we compare with a baseline called `Vanilla'\cite{yu2016sketch}, against which, our model only differs in fusing the final layer feature with the mid-layer feature. However, our model significantly outperforms the Vanilla model by a remarkable margin (around 11\% at acc.@1). This clearly demonstrates the effectiveness of the proposed framework and verifies the necessity of exploiting mid-level representation in FG-SBIR. (3)  Our model beats the attention-based model DSSA\footnote{Their full model contains three components, the attention module, coarse-fine fusion and HOLEF loss, which all contribute to the improvements.} \cite{song2017sketch} on all three datasets with a consistent margin around 2\% and outperforms their attention-only model by 10\%.  This is achieved with a simpler model architecture, suggesting the local discriminative feature encoded in the mid-level representation are more informative than the feature captured by the more complex attention module. 

\keypoint{Results on Person ReID} The results achieved on Market-1501, DukeMTMC\_reID and CUHK03-New are shown in Table~\ref{tab:market}, Table~\ref{tab:duke} and Table~\ref{tab:cuhk03} respectively. From the results we observe that: (1) The proposed model outperforms  state-of-the-art methods by a big margin on all  datasets. On Market-1501, it obtains an improvement of 7.6\%  and 3.1\% in Rank-1 accuracy and mAP respectively over the nearest competitor DaF \cite{yu2017divide}. On DukeMTMC\_reID, it achieves an improvement of 3.7\%  and 7.1\% in Rank-1 accuracy and mAP respectively over the strongest competitor SVDNet \cite{sun2017svdnet}. While on CUHK03-New \textit{Labelled} set, it outperforms the nearest competitor SVDNet \cite{sun2017svdnet} by 10.6\% and 9.7\% in Rank-1 and mAP, while on the \textit{Detected} set, it outperforms by 5.6\% and 6.2\% in Rank-1 and mAP.  (2) Although both HP-Net \cite{liu2017hydraplus} and SpindleNet \cite{zhao2017spindle} achieve large improvement by fusing features captured by an attention module or part detector(s) respectively, our proposed method outperforms them significantly with a much simpler architecture. (3) Again, to evaluate how much the introduced mid-level feature fusion contributes to our model's performance, we compare with the `Vanilla' model. This baseline uses the same base network, \ie, ResNet-50 but without fusing the final-layer feature with the mid-level representation. The results show that fusing the two types of features  can drastically improve the performance (on Market-1501, the margin in mAP is 8\% in SQ while 6.4\% in MQ; on DukeMTMC\_reID, the margins are 13.2\% and 7.8\% in mAP and Rank-1; on CUHK03-New, our model outperforms the Vanilla model by an average 11\% in mAP and Rank-1 on \textit{Labelled} set while around 10.7\% on the \textit{Detected} set). These improvements again confirm the efficacy of exploiting mid-level representations in CDIM.

\begin{table}[!htbp]
\normalsize
\centering
\resizebox{0.75\columnwidth}{!}{
\begin{tabular}{c|cc}
\hline
QMUL-Shoe & acc.@1  & acc.@10   \tabularnewline
\hline
HOG-BoW\cite{li2015free}+ rankSVM & 17.39\% & 67.83\%   \tabularnewline
Dense-HOG\cite{hu2013evalSKBIR} + rankSVM & 24.35\% & 65.22\%  \tabularnewline
ISN Deep\cite{qianyu2015bmvc} + rankSVM & 20.00\% & 62.61\%  \tabularnewline
DSSA(attention only)\cite{song2017sketch}  & 54.78\% & --  \tabularnewline
DSSA\cite{song2017sketch}  & 61.74\% & \textbf{94.78}\%  \tabularnewline
\hline
Vanilla\cite{yu2016sketch} & 52.17 \% & 92.17 \%  \tabularnewline
Ours & \textbf{64.35}\% & 93.04\%  \tabularnewline
\hline
\end{tabular}}

\resizebox{0.75\columnwidth}{!}{
\begin{tabular}{c|cc}
\hline
QMUL-Chair  & acc.@1  & acc.@10   \tabularnewline
\hline
BoW-HOG\cite{li2015free} + rankSVM & 28.87\% & 67.01\%  \tabularnewline
Dense-HOG\cite{hu2013evalSKBIR} + rankSVM & 52.57\% & 93.81\%  \tabularnewline
ISN Deep\cite{qianyu2015bmvc} + rankSVM & 47.42\% & 82.47\%  \tabularnewline
DSSA(attention only)\cite{song2017sketch} & 74.23\% & --  \tabularnewline
DSSA\cite{song2017sketch} & 81.44\% & 95.88\%  \tabularnewline
\hline
Vanilla\cite{yu2016sketch} & 72.16 \% & \textbf{98.96} \%  \tabularnewline
Ours &  \textbf{84.54}\% & 97.94\%   \tabularnewline
\hline
\end{tabular}}

\resizebox{0.75\columnwidth}{!}{
\begin{tabular}{c|cc}
\hline
QMUL-Handbag  & acc.@1  & acc.@10   \tabularnewline
\hline
HOG-BoW\cite{li2015free} + rankSVM & 2.38\% & 10.71\%   \tabularnewline
Dense-HOG\cite{hu2013evalSKBIR} + rankSVM & 15.47\% & 40.48\%  \tabularnewline
ISN Deep\cite{qianyu2015bmvc} + rankSVM & 9.52\% & 44.05\%  \tabularnewline
DSSA(attention only)\cite{song2017sketch} & 41.07\% & --  \tabularnewline
DSSA\cite{song2017sketch} & 49.40\% & 82.74\%  \tabularnewline
\hline
Vanilla\cite{yu2016sketch} &  39.88\% &  82.14\%  \tabularnewline
Ours & \textbf{50.00}\% & \textbf{83.33}\%  \tabularnewline
\hline
\end{tabular}}
\vspace{0.3cm}
\protect\caption{Comparative results on FG-SBIR.}
\centering
\label{tab:results_FG-SBIR}
\end{table}

\begin{table}[!htbp]
\normalsize
\centering
\resizebox{0.95\columnwidth}{!}{
\begin{tabular}{c|cc|cc}
\hline    
\multirow{2}{*}{Market-1501} &  \multicolumn{2}{c|}{single query} & \multicolumn{2}{c}{multi query}\\
\cline{2-5}
& mAP & Rank-1 & mAP & Rank-1\\ 
\hline
\ HP-net* \cite{liu2017hydraplus}  & -- & 76.9\% & -- & -- \\     
\ Spindle* \cite{zhao2017spindle}  & -- & 76.9\% & -- & -- \\   
\ Basel. + OIM \cite{xiao2017joint}  & -- & 82.1\% & -- & -- \\   
\ DPA* \cite{zhao2017deeply}  & 63.4\% & 81.0\% & -- & -- \\   
\ SVDNet \cite{sun2017svdnet}  & 62.1\% & 82.3\% & -- & -- \\   
\ DaF \cite{yu2017divide}  & 72.4\% & 82.3\% & -- & -- \\   
\ ACRN \cite{schumann2017person}   & 62.6\% & 83.6\% & -- & -- \\   
\ Context* \cite{li2017learning}  & 57.5\% & 80.3\% & 66.7\% & 86.8\% \\                           
\ JLML \cite{li2017person}  & 64.4\% & 83.9\% & 74.5\% & 89.7\% \\
\ Basel. + LSRO \cite{zheng2017unlabeled}  & 66.1\% & 84.0\% & 76.1\% &  88.4\% \\
\ SSM \cite{bai2017scalable}  & 68.8\% & 82.2\% & 76.2\% & 88.2\% \\
\hline
\ Vanilla & 67.55\% & 87.32\% & 75.98\% & 91.45\% \\
\ Ours & \textbf{75.55}\% &  \textbf{89.87}\% &  \textbf{82.37}\% & \textbf{93.32}\% \\
\hline
\end{tabular}}
\vspace{0.3cm}
\protect\caption{Comparative results on Market-1501. `*' indicates using attention module or part detector in the model.}
\centering
\label{tab:market}
\end{table}

\begin{table}[!htbp]
\normalsize
\centering
\resizebox{0.65\columnwidth}{!}{
\begin{tabular}{c|cc}
\hline
\ DukeMTMC\_reID & mAP & Rank-1 \\ 
\hline
\ Basel. + LSRO \cite{zheng2017unlabeled} & 47.13\% & 67.68\%  \\                           
\ Basel. + OIM \cite{xiao2017joint}  & -- & 68.1\%  \\
\ APR \cite{lin2017improving}  & 51.88\% & 70.69\%  \\
\ ACRN \cite{schumann2017person} & 51.96\% & 72.58\% \\
\ SVDNet \cite{sun2017svdnet}  & 56.8\% & 76.7\%  \\
\hline
\ Vanilla & 50.68\% & 72.67\% \\
\ Ours & \textbf{63.88}\% &  \textbf{80.43}\%  \\
\hline
\end{tabular}}
\vspace{0.3cm}
\protect\caption{Comparative results on DukeMTMC\_reID.}
\centering
\label{tab:duke}
\end{table}

\begin{table}[!htbp]
\normalsize
\centering
\resizebox{0.85\columnwidth}{!}{
\begin{tabular}{c|cc|cc}
\hline    
\multirow{2}{*}{CUHK03-New} &  \multicolumn{2}{c|}{Labelled} & \multicolumn{2}{c}{Detected}\\
\cline{2-5}
& mAP & Rank-1 & mAP & Rank-1\\ 
\hline
\ DaF\cite{yu2017divide}   & 31.5\% & 27.5\% & 30.0\% & 26.4\% \\     
\ SVDNet\cite{sun2017svdnet}  & 37.8\% & 40.9\% & 37.3\% & 41.5\% \\   
\ Re-rank\cite{zhong2017re} & 40.3\% & 38.1\% & 37.4\% & 34.7\% \\
\hline
\ Vanilla & 35.77\% & 40.64\% & 32.82\% & 36.36\% \\
\ Ours & \textbf{47.52}\% &  \textbf{51.50}\% &  \textbf{43.51}\% & \textbf{47.14}\% \\
\hline
\end{tabular}}
\vspace{0.3cm}
\protect\caption{Comparative results on CUHK03-New.}
\centering
\label{tab:cuhk03}
\end{table}

\subsection{Further Analysis}\label{sec:discuss}

\noindent\textbf{Which Layer(s) to Fuse} \quad As discussed in Sec.~\ref{sec:alternatives}, we select the layers to fuse motivated by the findings in \cite{bau2017network,diba2016deepCamp,vittayakorn2016neuralAttrDiscovery}. Other layer selections exist. In this section, we compare the effect of fusing different layers to provide some insights. Specifically, for FG-SBIR, we explore the activations of \textit{conv3},  \textit{conv4} and a combination of \textit{conv4} and \textit{conv5}. The results are listed in Table \ref{tab:fusion}. From the results, it is clear to see that the middle layer \textit{conv5} used in our model is more informative than the earlier layers. 

Similar experiments are carried out on person ReID. For ResNet-50 which has 4 residual blocks, there are more fusion options and here we compare ours with the following alternatives: (1) the 6 units in block3, \ie \textit{res4x}, $x\in\{a,b,c,d,e,f\}$, (2) block 1, 2, 3, \ie \textit{res2} \textit{res3} and \textit{res4}, (3) block 1, 2, 3 and \textit{res5a} and \textit{res5b} (this is actually a combination of our selected fusion layers and all previous residual blocks). Additionally, we also compare with a `Vanilla+fusion' model which directly fuses the features extracted from the same layers (as we used in our proposed model) of a well-trained ResNet-50 (\ie the baseline model `Vanilla'). From the results in Table~\ref{tab:fusion layer}, we find that (a) each layer contains some useful information complementary to the top-layer high-level representation (all options improve on Vanilla); (b) activations of the \textit{res5a} and \textit{res5b} are more informative than other layers; and (c) it is interesting to see that fusing all blocks according to (3) gives inferior performance than ours. This indicates that some low-level features can be redundant.
These observations verify the idea that different layers contain different information useful for CDIM, and validate our specific design choices. However fully automated selection of the informativeness of each layer is still an open question to be addressed in future work. 


\begin{table}[!t]
\normalsize
\centering
\resizebox{0.55\columnwidth}{!}{
\begin{tabular}{c|cc}
\hline
Layer(s) & acc.@1  & acc.@10  \tabularnewline
\hline
conv3 & 60.0\% & 92.17\%  \tabularnewline
conv4 & 60.87\% & 90.43\%  \tabularnewline
conv4,5 & 62.61\% & 93.04\% \tabularnewline
conv5 (Ours) & \textbf{64.35}\% & \textbf{93.04}\%  \tabularnewline
\hline
\end{tabular}}
\vspace{0.3cm}
\protect\caption{Comparison of fusing different layers for FG-SBIR on QMUL-Shoe.}
\centering
\label{tab:fusion}
\end{table}

\begin{table}[!t]
\normalsize
\centering
\resizebox{1.0\columnwidth}{!}{
\begin{tabular}{c|cc|cc}
\hline    
\multirow{2}{*}{Models} &  \multicolumn{2}{c|}{single query} & \multicolumn{2}{c}{multi query}\\
\cline{2-5}
& mAP & Rank-1 & mAP & Rank-1\\ 
\hline
\ Vanilla+fusion  & 71.88\% & 88.87\% & 79.64\% & 92.70\% \\                       
\ res2, res3, res4  & 71.24\% & 88.36\% & 79.02\% & 92.43\% \\
\ res4(a-f)  & 73.17\% & 89.90\% & 80.58\% &  \textbf{93.53}\% \\     
\ res2,3,4 + res5(a,b)   & 74.52\% & \textbf{89.93}\% & 81.85\% & 93.00\% \\
\ res5(a,b) (Ours)  & \textbf{75.55}\% &  89.87\% &  \textbf{82.37}\% & 93.32\% \\
\hline
\end{tabular}}
\vspace{0.3cm}
\protect\caption{Results of fusing different layers for ReID on Market-1501.}
\centering
\label{tab:fusion layer}
\end{table}

\begin{figure*}[!ht]
\centering
\includegraphics[width=\linewidth]{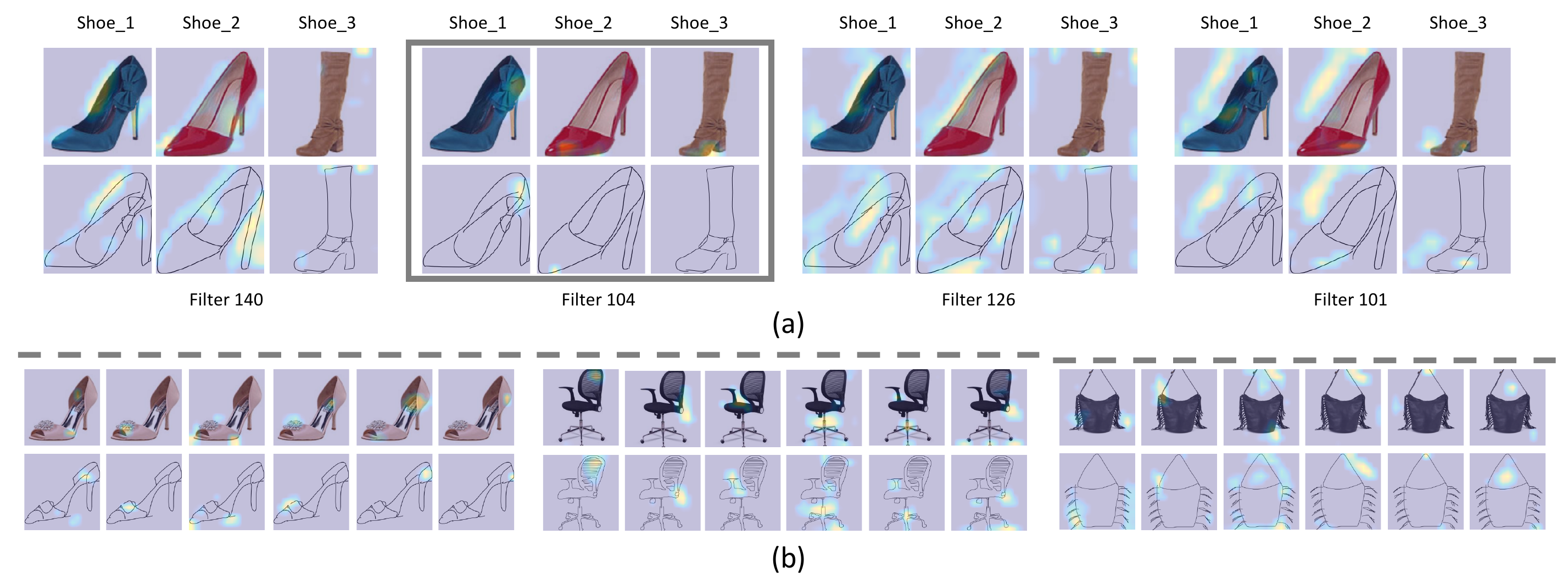}
\vspace{-0.6cm}
\caption{Visualizations of the feature maps activated by the filters in layer \textit{conv5} of our  FG-SBIR model. Each column are a sketch-photo pair. (a) Visualizations of a triplet pair activated by 4 different filters. (b) Visualizations of the feature map activated by different filters on the same sketch-photo pairs. }
\label{fig:triplet}
\end{figure*}

\keypoint{Comparison of Pooling Strategies} The pooling strategy is another factor affecting the performance of the fusion module.
In this experiment, we evaluate the impact of the pooling strategy on the final model performance. As shown in Table~\ref{tab:pooling FG-SBIR} and Table~\ref{tab:pooling reid}, the results suggests that for a pose-aligned task such as FG-SBIR, spatial information is indeed important and flattening is the correct choice. On the contrary,  spatial information is a distractor in ReID where person pose and viewpoint can change drastically across views resulting in large mis-alignment. 

\begin{table}[!htbp]
\normalsize
\centering
\resizebox{0.55\columnwidth}{!}{
\begin{tabular}{c|cc}
\hline
Pooling & acc.@1  & acc.@10   \tabularnewline
\hline
GAP & 58.26\% & 93.04\%  \tabularnewline
Flattening (Ours) & \textbf{64.35}\% & 93.04\%  \tabularnewline
\hline
\end{tabular}}
\vspace{0.3cm}
\protect\caption{Comparison of different pooling methods for FG-SBIR on QMUL-Shoe.}
\centering
\label{tab:pooling FG-SBIR}
\end{table}

\begin{table}[!htbp]
\normalsize
\centering
\resizebox{0.8\columnwidth}{!}{
\begin{tabular}{c|cc|cc}
\hline    
\multirow{2}{*}{Pooling} &  \multicolumn{2}{c|}{single query} & \multicolumn{2}{c}{multi query}\\
\cline{2-5}
& mAP & Rank-1 & mAP & Rank-1\\ 
\hline
\ Flattening & 71.58\% & 89.25\% & 78.56\% & 92.28\% \\
\ GAP (Ours) & \textbf{75.55}\% &  \textbf{89.87}\% &  \textbf{82.37}\% &  \textbf{93.32}\% \\
\hline
\end{tabular}}
\vspace{0.3cm}
\protect\caption{Comparison of different pooling methods for ReID on Market-1501.}
\centering
\label{tab:pooling reid}
\end{table}

\begin{table}[!htbp]
\normalsize
\centering
\resizebox{0.75\columnwidth}{!}{
\begin{tabular}{c|cc}
\hline
QMUL-Shoe & acc.@1  & acc.@10   \tabularnewline
\hline
InceptionV3 & 57.39\% & 91.30\%  \tabularnewline
InceptionV3+fusion module & \textbf{72.12}\% & \textbf{96.52}\%  \tabularnewline
\hline
\end{tabular}}

\resizebox{0.75\columnwidth}{!}{
\begin{tabular}{c|cc}
\hline
QMUL-Chair  & acc.@1  & acc.@10   \tabularnewline
\hline
InceptionV3 & 88.66\% & 98.97\%  \tabularnewline
InceptionV3+fusion module & \textbf{97.94}\% & \textbf{100.0}\%  \tabularnewline
\hline
\end{tabular}}

\resizebox{0.75\columnwidth}{!}{
\begin{tabular}{c|cc}
\hline
QMUL-Handbag  & acc.@1  & acc.@10   \tabularnewline
\hline
InceptionV3 & 45.83\% & 82.14\%  \tabularnewline
InceptionV3+fusion module & \textbf{65.48}\% & \textbf{90.48}\%  \tabularnewline
\hline
\end{tabular}}
\vspace{0.3cm}
\protect\caption{When the fusion module works with InceptionV3.}
\centering
\label{tab:generalise}
\end{table}

\keypoint{Evaluation on the Applicability of the Approach }
We have  demonstrated that our proposed design pattern can be applied to two different CDIM problems with different base CNN and loss functions. In this experiment, we further show that for the same problem, the mid-level feature fusion strategy also boosts performance with different base networks.  Here we replace the Sketch-a-Net with InceptionV3 \cite{szegedy2016inception} in the  FG-SBIR task. 
We fine-tune it based on the ImageNet pre-trained model.  We choose the layer \textit{Mixed\_7b} for fusion.
Table~\ref{tab:generalise} shows that (1) with this model, even better performance is obtained on all three datasets. This is not surprising since InceptionV3 is much bigger and deeper. (2) The gap between the model with and without mid-level feature fusion remains big, confirming that the proposed approach can be applied to different CNN base network for the same problem. 

\noindent\textbf{Qualitative Results} \quad We visualize the feature maps activated by the filters in layer \textit{conv5} of our FG-SBIR model.  Fig.~\ref{fig:triplet}(a) compares the feature maps of three photo-sketch pairs. It is clear to see that Shoe\_1 and Shoe\_2, which are visually similar, also have similar activations. Apart from color, the most obvious difference between Shoe\_1 and Shoe\_2 is the bowknot. `Filter 104' detects the `bowknot' on Shoe\_1 but does not find it on Shoe\_2. Instead, it activates a different part. And Filter 104 also does not have activation on Shoe\_3. In (b), we show different filter activations on the same photo-sketch pairs. We observe that: (a) the feature captured by the filters of layer \textit{conv5} are domain-invariant; (b) some filters may detect similar parts/regions, \eg, the last two filters of the shoe example both detect the back region. This may suggest that some filters are redundant and can be further pruned; and (c) some filters posses the ability of detecting multiple parts, \ie the part structure, \eg, in the chair example, the second filter from the right detects both the arm and the base of the chair.

\vspace{-0.3cm}
\section{Conclusions}
\vspace{-0.2cm}
We  proposed a simple deep architecture design pattern for cross-domain instance matching (CDIM). The key idea is based on fusing mid-level features with high-level features to capture discriminative  features of different semantic levels. We argued that the mid-level features already exist in the mid-layer feature maps which just need to be represented properly and fused with the final-layer features. Extensive experiments on both FG-SBIR and person ReID validated the effectiveness of the proposed approach.


{\small
\bibliographystyle{ieee}
\bibliography{CDIM_Qian_arxiv_v3}
}

\end{document}